%% file: main.tex
\documentclass{article} % For LaTeX2e
\usepackage{iclr2023_conference,times}

\input{math_commands.tex}

\usepackage{hyperref}
\usepackage{url}

\usepackage{wrapfig}
\usepackage{comment}

\usepackage{nicefrac}       % compact symbols for 1/2, etc.
\usepackage{microtype}      % microtypography
\usepackage{xcolor}         % colors
\usepackage{subcaption}
\usepackage{multirow}
\usepackage{verbatim}
\usepackage{diagbox}
\usepackage{bm}
\usepackage{makecell}
\usepackage{graphicx}
\usepackage{multirow}
\usepackage{amsmath}
\usepackage[english]{babel}
\usepackage{amsthm}
\usepackage{mathtools}
\usepackage[ruled,vlined]{algorithm2e}
\usepackage{xcolor}
\usepackage[utf8]{inputenc}
\usepackage{csquotes}
\usepackage{enumitem}
\usepackage{soul}
\definecolor{commentcolor}{RGB}{110,154,155}   % define comment color
\definecolor{operatorcolor}{RGB}{240,95,129} 
\newcommand{\PyComment}[1]{\ttfamily\textcolor{commentcolor}{\# #1}}  % add a "#" before the input text "#1"

\newcommand{\Pyoperator}[1]{\ttfamily\textcolor{operatorcolor}{#1}} % \ttfamily is the code font

\title{ESD: Expected Squared Difference as a Tuning-Free Trainable Calibration Measure}

\author{Hee Suk Yoon$^{1}$\thanks{Equal contribution}\,\,\,\,\,\,\,\,\,\, Joshua Tian Jin Tee$^{1}$\footnotemark[1]\,\,\,\,\,\,\,\,\,\, Eunseop Yoon$^{1}$\,\,\,\,\,\,\,\,\,\, Sunjae Yoon$^{1}$\,\,\,\,\,\,\,\,\,\, \\\bf{Gwangsu Kim}$^{1}$\,\,\,\,\,\,\,\,\,\, \bf{Yingzhen Li}$^{2}$\,\,\,\,\,\,\,\,\,\, \bf{Chang D. Yoo}$^{1}$\thanks{Corresponding Author} \\
$^{1}$Korea Advanced Institute of Science and Technology (KAIST)\,\,\,\,\,\,\,\,\,\,$^{2}$Imperial College London\\
\texttt{\{hskyoon,joshuateetj,esyoon97,sunjae.yoon\}@kaist.ac.kr} \\
\texttt{s88012@gmail.com}\,\,\,\,\,\,\,\texttt{yingzhen.li@imperial.ac.uk}\,\,\,\,\,\,\,\texttt{cd\_yoo@kaist.ac.kr}  \\ 
}
\iclrfinalcopy % Uncomment for camera-ready version, but NOT for submission.
\begin{document}

\maketitle

\begin{abstract}
Studies have shown that modern neural networks tend to be poorly calibrated due to over-confident predictions. Traditionally, post-processing methods have been used to calibrate the model after training. In recent years, various trainable calibration measures have been proposed to incorporate them directly into the training process. %However, these methods all incorporate internal hyperparameters introduced in the process of obtaining a differential calibration measure. Consequently, the performance of these calibration objectives relies on tuning these hyperparameters, incurring more computational cost as the size of neural networks and datasets become larger.
However, these methods all incorporate internal hyperparameters, and the performance of these calibration objectives relies on tuning these hyperparameters, incurring more computational costs as the size of neural networks and datasets become larger.
As such, we present Expected Squared Difference (ESD), a tuning-free (i.e., hyperparameter-free) trainable calibration objective loss, where we view the calibration error from the perspective of the squared difference between the two expectations. With extensive experiments on several architectures (CNNs, Transformers) and datasets, we demonstrate that (1) incorporating ESD into the training improves model calibration in various batch size settings without the need for internal hyperparameter tuning, (2) ESD yields the best-calibrated results compared with previous approaches, and (3) ESD drastically improves the computational costs required for calibration during training due to the absence of internal hyperparameter. The code is publicly accessible at \href{https://github.com/hee-suk-yoon/ESD}{https://github.com/hee-suk-yoon/ESD}.
%another way of defining calibration error that fixes these issues.
\end{abstract}
\section{Introduction}
The calibration of a neural network measures the extent to which its predictions align with the true probability distribution. Possessing this property becomes especially important in real-world applications, such as identification \citep{identification1, identification2}, autonomous driving \citep{autodriving, automousdriving2}, and medical diagnosis \citep{medical1, medical2}, where uncertainty-based decisions of the neural network are crucial to guarantee the safety of the users. However, despite the success of modern neural networks in accurate classification, they are shown to be poorly calibrated due to the tendency of the network to make predictions with high confidence regardless of the input. (i.e., over-confident predictions) \citep{calibrationmodern}.

Traditionally, post-processing methods have been used, such as temperature scaling and vector scaling \citep{calibrationmodern}, to calibrate the model using the validation set after the training by adjusting the logits before the final softmax layer. Various trainable calibration objectives have been proposed recently, such as MMCE \citep{MMCE} and SB-ECE \citep{softcal}, which are added to the loss function as a regularizer to jointly optimize accuracy and calibration during training. A key advantage of calibration during training is that it is possible to cascade post-processing calibration methods after training to achieve even better-calibrated models. Unfortunately, these existing approaches introduce additional hyperparameters in their proposed calibration objectives, and the performance of the calibration objectives is highly sensitive to these design choices. Therefore these hyperparameters need to be tuned carefully on a per model per dataset basis, which greatly reduces their viability for training on large models and datasets. 

%However, in the process of obtaining trainable calibration measures, all previously proposed methods introduce hyperparameters within the calibration objective loss. For example, MMCE uses the kernel embedding method to obtain a trainable calibration objective loss, in which it introduces the kernel bandwidth hyperparameter. In addition, SB-ECE also introduces a hyperparameter when softening the bins in the Expected Calibration Error (ECE) \citep{ECE}, which is a metric widely used to quantify calibration error by binning model predictions based on their confidence. These hyperparmaters need to be tuned since the optimal hyperparameters could change between different models and datasets (Figure \ref{fig:1}), which greatly reduces its viability for training with large models and datasets. 

To this end, we propose Expected Squared Difference (ESD), a trainable calibration objective loss that is hyperparameter-free. ESD is inspired by the KS-Error \citep{kserror}, and it views the calibration error from the perspective of the difference between the two expectations. 
In detail, our contributions can be summarized as follows:

\begin{itemize}
    \item We propose ESD as a trainable calibration objective loss that can be jointly optimized with the negative log-likelihood loss (NLL) during training. ESD is a binning-free calibration objective loss, and no additional hyperparameters are required. We also provide an unbiased and consistent estimator of the Expected Squared Difference, and show that it can be utilized in small batch train settings. 
    
    \item With extensive experiments, we demonstrate that across various architectures (CNNs \& Transformers) and datasets (in vision \& NLP domains), ESD provides the best calibration results when compared to previous approaches. The calibrations of these models are further improved by post-processing methods. 
    
    \item We show that due to the absence of an internal hyperparameter in ESD that needs to be tuned, it offers a drastic improvement compared to previous calibration objective losses with regard to the total computational cost for training. The discrepancy in computational cost between ESD and tuning-required calibration objective losses becomes larger as the model complexity and dataset size increases. 
\end{itemize}

\section{Related Work}
\label{headings}
Calibration of neural networks has gained much attention following the observation from \citet{calibrationmodern} that modern neural networks are poorly calibrated. One way to achieve better calibration is to design better neural network architectures tailored for uncertainty estimation, e.g., Bayesian Neural Networks \citep{blundell2015weight,gal2016dropout} and Deep Ensembles \citep{lakshminarayanan2017simple}. Besides model design, \textbf{post-processing} calibration strategies have been widely used to calibrate a trained machine learning model using a hold-out validation dataset. Examples include temperature scaling \citep{calibrationmodern}, which scales the logit output of a classifier with a temperature parameter; Platt scaling \citep{platt}, which fits a logistic regression model on top of the logits; and Conformal prediction \citep{vovk2005algorithmic,lei2018distribution} which uses validation set to estimate the quantiles of a given scoring function. Other post-processing techniques include histogram binning \citep{histogram}, isotonic regression \citep{isotonic}, and Bayesian binning into quantiles \citep{BBQ}. 

Our work focuses on \textbf{trainable calibration methods}, which train neural networks using a hybrid objective, combining a primary training loss with an auxiliary calibration objective loss. In this regard, one popular objective is Maximum Mean Calibration Error (MMCE) \citep{MMCE}, which is a kernel embedding-based measure of calibration that is differentiable and, therefore, suitable as a calibration loss. Moreover, \citet{softcal} proposes a trainable calibration objective loss, SB-ECE, and S-AvUC, which softens previously defined calibration measures.

\section{Problem Setup}
\label{headings}
%This section introduces the notion of calibration error in neural networks and the metric widely used to estimate it. We then introduce the framework and pre-existing problems of calibration during training. The notations defined in this section will be used throughout the paper.
\subsection{Calibration Error and Metric}
Let us first consider an arbitrary neural network as $f_\theta: \mathcal{D} \rightarrow [0,1]^C$ with network parameters $\theta$, where $\mathcal{D}$ is the input domain and $C$ is the number of classes in the multiclass classification task. Furthermore, we assume that the training data, $(\boldsymbol{x_i},\boldsymbol{y_i})_{i=1}^n$ are sampled {\it i.i.d.} from the joint distribution $\mathbb{P}(\boldsymbol{X},\boldsymbol{Y})$ (here we use one-hot vector for $\boldsymbol{y}$). Here, $\boldsymbol{Y} = (Y_1,...,Y_C)$, and $\boldsymbol{y} = (y_1,...,y_C)$ are samples from this distribution. We can further define a multivariate random variable $\boldsymbol{Z} = f_\theta(\boldsymbol{X})$ as the distribution of the outputs of the neural network. Similarly, $\boldsymbol{Z} = (Z_1,...,Z_C)$, and $\boldsymbol{z} = (z_1,...,z_C)$ are samples from this distribution.  We use $(z_{K,i},y_{K,i})$ to denote the output confidence and the one-hot vector element associated with the $K$-th class of the $i$-th training sample. Using this formulation, a neural network is said to be perfectly calibrated for class $K,$ if and only if
\begin{eqnarray} \label{eq:1}
\mathbb{P}(Y_K = 1 | Z_K = z_{K}) = z_K.
\end{eqnarray}
Intuitively, Eq. (\ref{eq:1}) requires the model accuracy for class $K$ to be $z_K$ on average for the inputs where the neural network produces prediction of class $K$ with confidence $z_K$. In many cases, the research of calibration mainly focuses on the calibration of the max output class that the model predicts. Thus, calibration error is normally reported with respect to the predicted class (i.e., max output class) only. As such, $k$ will denote the class with maximum output probability. Furthermore, we also write $I(\cdot)$ as the indicator function which returns one if the Boolean expression is true and zero otherwise.

With these notations, one common measurement of calibration error is the difference between confidence and accuracy which is mathematically represented as \citep{calibrationmodern}
\begin{eqnarray}
\mathbb{E}_{Z_k}\left[ \vert \mathbb{P}(Y_k = 1 | Z_k)-Z_k \vert \right].
\end{eqnarray}

To estimate this, the Expected Calibration Error (ECE) \citep{ECE} uses $B$ number of bins with disjoint intervals $B_j = (\frac{j}{B},\frac{j+1}{B}]$, $j = 0, 1 , ..., B-1$ to compute the calibration error as follows:
\begin{eqnarray}
\text{ECE}  = \frac{1}{|\mathcal{D}|} \sum^{B-1}_{j=0}\sum^N_{i=1} \left\vert I \left(\frac{j}{B}< z_{k,i}\leq \frac{j+1}{B}, y_{k,i}= 1 \right)  - z_{k,i}) \right\vert.
\end{eqnarray}
\subsection{Calibration During Training}
Post-processing method and calibration during training are the two primary approaches for calibrating a neural network. Our focus in this paper is on the latter, where our goal is to train a calibrated yet accurate classifier directly. Note that calibration and predictive accuracy are independent properties of a classifier. In other words, being calibrated does not imply that the classifier has good accuracy and vice versa. Thus, training a classifier to have both high accuracy and good calibration requires jointly optimizing a calibration objective loss alongside the negative log-likelihood (NLL) with a scaling parameter $\lambda$ for the secondary objective:
\begin{equation}
\underset{\theta}{\min} \ \text{NLL}(\mathcal{D},\theta)+\lambda \cdot \text{CalibrationObjective}(\mathcal{D},\theta).
\end{equation}
\subsubsection{Existing Trainable Calibration Objectives need Tuning}
\citet{MMCE} and \citet{softcal} suggested that the disjoint bins in ECE can introduce discontinuities which are problematic when using it as a calibration loss in training. Therefore, additional parameters were introduced for the purpose of calibration during training.
%Various trainable calibration objectives have been proposed in the past. 
For example, \citet{MMCE} proposed Maximum Mean
Calibration Error (MMCE), where it utilizes a Laplacian Kernel instead of the disjoint bins in ECE. Furthermore, \citet{softcal} proposed soft-binned ECE (SB-ECE) and soft AvUC, which are softened versions of the ECE metric and the AvUC loss \citep{avuc}, respectively. All these approaches address the issue of discontinuity which makes the training objective differentiable. However they all introduce additional design choices - MMCE requires a careful selection of the kernel width $\phi$, SB-ECE needs to choose the number of bins $M$ and a softening parameter $T$, and S-AvUC requires a user-specified entropy threshold $\kappa$ in addition to the softening parameter $T$. Searching for the optimal hyperparameters can be computationally expensive especially as the size of models and dataset become larger. 
\begin{figure}[t]
	\centering
	\includegraphics[width=\linewidth]{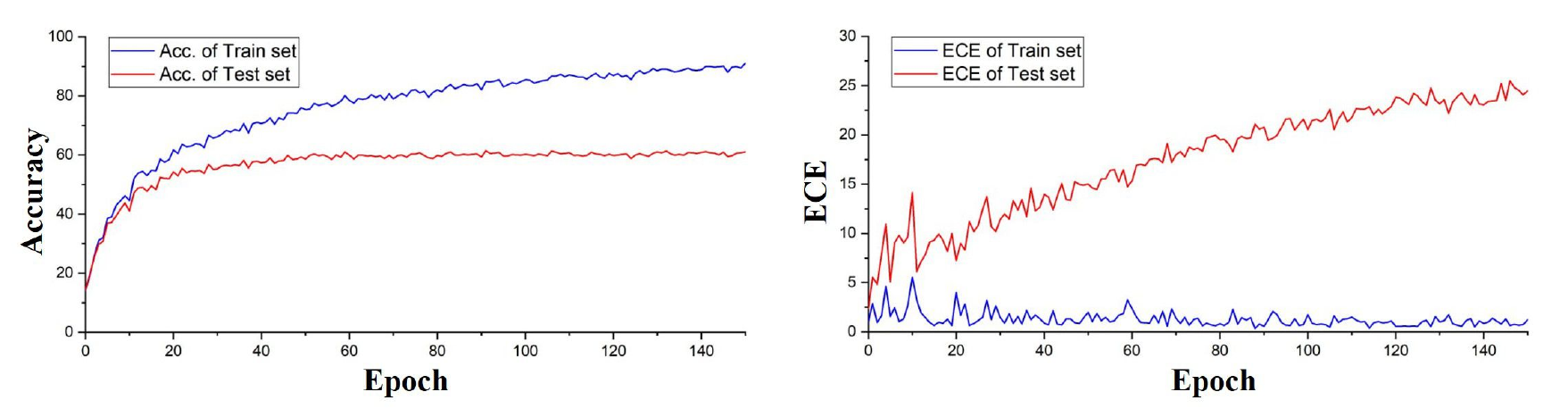}
	\caption{Accuracy (\%) curve (left) and its corresponding ECE (\%) curve (right) during training with negative log-likelihood (NLL) loss. It could be seen that since NLL implicitly trains for calibration error, the ECE of the train set approaches zero while the ECE of the test set increases during training.}
	\label{fig:2}
\end{figure}
\subsubsection{Calibration During Training Suffers from Over-fitting Problem}
\label{overfitproblem}
NLL loss is known to implicitly train for calibration since it is a proper scoring rule, so models trained with NLL can overfit in terms of calibration error \citep{focal, softcal}. Figure \ref{fig:2} provides an example of an accuracy curve and its corresponding ECE curve for a model trained with NLL loss. We see that the model is overfitting in terms of both accuracy and ECE, which causes the gap between train and test ECE to become larger during training. As such, adding a calibration objective to the total loss will not be able to improve model calibration as it mainly helps improve calibration of the model with respect to the training data only.  

\citet{MMCE} tackle this issue by introducing a weighted version of their calibration objective loss. They consider using larger weights to incorrect prediction samples after observing that the fraction of incorrect to the correct samples on the training data is smaller than that of the validation and test data. Instead of changing the calibration objective function itself, \citet{softcal} introduced a new training scheme called \textit{interleaved training}, where they split the training data further and dedicate a small portion of it to optimize the calibration objective loss. We will follow this strategy and introduce the technical details in the later sections.

\section{ESD: Expected Squared Difference}

We propose Expected Squared Difference (ESD) as a tuning-free (i.e., hyperparameter-free) calibration objective. Our approach is inspired by the viewpoint of calibration error as a measure of distance between two distributions to obtain a binning-free calibration metric \citep{kserror}. %This allows us to avoid having to introduce kernels or softening operations to tackle the bins internal to the calibration metrics, such as ECE, to make it suitable for training.
By using this approach, there is no need to employ kernels or softening operations to handle the bins within calibration metrics, such as ECE, in order to make them suitable for training.

In particular, we consider calibration error as the difference between the two expectations. We first start with the definition of perfect calibration in Eq. (\ref{eq:1}):
\begin{equation}
\begin{aligned}
   & \mathbb{P}(Y_k = 1 | Z_k = z_{k}) = z_k && \forall z_k\in [0,1]\\
\Leftrightarrow \ \ &\mathbb{P}(Y_k = 1, Z_k = z_k) = z_k\mathbb{P}(Z_k = z_k)   &&\forall z_k\in [0,1] &&&\text{(by Bayes rule)}.
\end{aligned}
\end{equation}

Now considering the accumulation of the terms on both sides for arbitrary confidence level $\alpha \in [0, 1]$, the perfect calibration for an arbitrary class $k$ can now be written as:
\begin{equation}
\begin{aligned}
\int^\alpha_{0}\mathbb{P}(Y_k = 1, Z_k = z_k)dz_k &= \int^\alpha_{0}z_k\mathbb{P}(Z_k = z_k)dz_k\\
\Leftrightarrow \mathbb{E}_{Z_k,Y_k}[I(Z_{k}\leq \alpha,Y_k=1)] &= \mathbb{E}_{Z_k,Y_k}[Z_{k}I(Z_{k}\leq \alpha)],  &&\forall \alpha\in [0,1].
\end{aligned}
\end{equation}
This allows us to write the difference between the two expectations as
\begin{equation}
\label{eq:7}
\begin{aligned}
d_k(\alpha) &= \left\vert \int^\alpha_{0}\mathbb{P}(Y_k = 1, Z_k = z_k) - z_k\mathbb{P}(Z_k = z_k)dz_k \right\vert\\
            &= |\mathbb{E}_{Z_k,Y_k}[I(Z_{k}\leq \alpha)(I(Y_k=1)-Z_{k})]|,
\end{aligned}
\end{equation}
and $d_k(\alpha) = 0$, $\forall \alpha\in [0,1]$ if and only if the model is perfectly calibrated for class $k$. Since this has to hold $\forall \alpha\in [0,1]$, we propose ESD as the expected squared difference between the two expectations:
\begin{equation}
\label{eq:esd_definition}
\mathbb{E}_{Z'_k}[d_k(Z'_k)^2] = \mathbb{E}_{Z'_k}[\mathbb{E}^2_{Z_k,Y_k}[I(Z_{k}\leq Z^{'}_{k})(I(Y_k=1)-Z_{k})]].
\end{equation}
Due to the close relationship between $d_k(\alpha)$ and calibration of a neural network, the difference between an uncalibrated and a calibrated neural network can be clearly observed using $\mathbb{E}_{Z'_k}[d_k(Z'_k)^2]$ as visually shown in Appendix \ref{intuition}. Since ESD = 0 iff. the model is perfectly calibrated as shown in the following theorem, this metric is a good measure of calibration. 

{\bf Theorem 1.}
{\it 
$\mathbb{E}_{Z'_k}[\mathbb{E}^2_{Z_k,Y_k}[I(Z_{k}\leq Z^{'}_{k})(I(Y_k=1)-Z_{k})] ] = 0$ iff.~the model is perfectly calibrated.
}
\begin{proof} 
Since $d_k(Z'_k)^2 = \mathbb{E}^2_{Z_k,Y_k}[I(Z_{k}\leq Z'_{k})(I(Y_k=1)-Z_{k})]$ is a non-negative random variable induced by $Z'_k$, \  $ \mathbb{E}_{Z'_k}[d_k(Z'_k)^2] = 0 $ iff. $ 
\mathbb{P}(d_k(Z'_k) =0) =1.$ Furthermore, 
\begin{eqnarray*}
\mathbb{P}(d_k(Z'_k) =0) =1  
&\iff & d_k(\alpha) =0 \  \hspace{2mm}\forall \alpha \in \mathcal{I}\text{ where $\mathcal{I}$ is the support set of $Z'_k$ }.  
\end{eqnarray*}
Thus, $$ \mathbb{E}_{Z'_k}[d_k(Z'_k)^2] = 0 \text{ iff. } d_k(\alpha) =0 \  \hspace{5mm}\forall \alpha \in \mathcal{I}.$$ 
From lemma 1.1 below, we have $ d_k ( \alpha ) = 0\hspace{2mm} \forall \alpha \in [0,1]$ iff. $d_k(\alpha) =0 \  \hspace{2mm}\forall \alpha \text{ s.t. } \mathbb{P}(Z'_k=\alpha) \neq 0.$ Consequently, $\mathbb{E}_{Z'_k}[\mathbb{E}^2_{Z_k,Y_k}[I(Z_{k}\leq Z'_{k})(I(Y_k=1)-Z_{k})]] = 0$ iff.~the model is perfectly calibrated.   
%&\iff& d_k(\alpha) =0 \ \hspace{5mm}\forall \alpha	\in[0,1] \hspace{17.5mm}\text{(From lemma 1.1 below)}
% $$\end{eqnarray*}
%Thus, $\mathbb{E}_{Z'_k}[\mathbb{E}^2_{Z_k,Y_k}[I(Z_{k}\leq Z'_{k})(I(Y_k=1)-Z_{k})]] = 0$ iff.~the model is perfectly calibrated.
\end{proof}
{\bf Lemma 1.1}
{\it Let $\mathcal{I}$ be the support set of random variable $Z_k$, then $d_k(\alpha) = 0$ $ \forall \alpha \in \mathcal{I}$ iff. $d_k(\alpha) = 0 $ $\forall \alpha	\in[0,1]$.}
\begin{proof}
The backward direction result is straight-forward, so we only prove for the forward direction: If $d_k(\alpha) = 0$ $ \forall \alpha \in \mathcal{I}$, then $d_k(\alpha) = 0 $ $\forall \alpha	\in[0,1]$.\\ 
For arbitrary $\alpha' \in \mathcal{I}^c$, let $\alpha = \argmin_{a \in \mathcal{I}} \vert a  - \alpha'|$ and $\alpha<\alpha'$ .  We then have,
\begin{equation}
\begin{aligned}
d_k(\alpha') &= |\int_0^{\alpha'}\mathbb{P}(Y_k =1,Z_k = z_k)-z_k \mathbb{P}(Z_k=z_k)dz_k|
\\
		     &= |\int_0^{\alpha'}\mathbb{P}(Y_k=1|Z_k = z_k)\mathbb{P}(Z_k=z_k)-z_k \mathbb{P}(Z_k=z_k)dz_k|\\
             &=  |\int_0^{\alpha}\mathbb{P}(Y_k=1|Z_k = z_k)\mathbb{P}(Z_k=z_k)-z_k \mathbb{P}(Z_k=z_k)dz_k \\
             &  \hspace{4mm}  + \int_\alpha^{\alpha'}\mathbb{P}(Y_k=1|Z_k = z_k)\mathbb{P}(Z_k=z_k)-z_k \mathbb{P}(Z_k=z_k)dz_k|   % \vert \text{ is nearest } \alpha \in \mathcal{I} \text{ to } \alpha’\\
             \\
             &=  |\int_0^\alpha \mathbb{P}(Y_k=1|Z_k = z_k)\mathbb{P}(Z_k=z_k)-z_k \mathbb{P}(Z_k=z_k)dz_k| \\
             &= 0.
             \notag 
\end{aligned}
\end{equation}
\end{proof}
\subsection{An estimator for ESD}
In this section, we use $(z_{k,i},y_{k,i})$ to denote the output confidence and the one-hot vector element associated with the $k$-th class of the $i$-th training sample respectively. As the expectations in the true Expected Squared Difference (Eq. (\ref{eq:esd_definition})) are intractable, we propose a Monte Carlo estimator for it which is unbiased. A common approach is to use a naive Monte Carlo sampling with respect to both the inner and outer expectations to give the following:
\begin{eqnarray}
%\begin{split}
&&\mathbb{E}_{Z'_k}[\mathbb{E}^2_{Z_k,Y_k}[I(Z_{k}\leq Z^{'}_{k})(I(Y_k=1)-Z_{k})]] \approx \frac{1}{N}\sum^{N}_{i=1}\bar{g_i}^2,\hspace{7mm} \label{eq:10}  \\ 
&&\text{ where }\bar{g_i} = {\frac{1}{N-1}}\sum^N\limits_{\substack{j=1\\j\neq i}}g_{ij}\text{ and }
g_{ij} = I(z_{k,j}\leq z_{k,i})[I(y_j=k)-z_{k,j}]. \notag 
%\end{split}
\end{eqnarray}
However, Eq. (\ref{eq:10}) results in a biased estimator that is an upper bound  of the true Expected Squared Difference. To account for the bias, we propose the unbiased and consistent estimator (proof in Appendix \ref{unbias}) of the true Expected Squared Difference\footnote{To avoid confusion, ESD from this point onward will refer to the estimator instead of its expectation form.},
\begin{equation}
\begin{aligned}
ESD = \frac{1}{N}\sum^{N}_{i=1}\left[\bar{g_i}^2-\frac{S_{g_i}^2}{N-1}\right]
\text{\hspace{2mm}where\hspace{1mm}}
S_{g_i}^2 = \frac{1}{N-2}\sum^{N}_{\substack{j=1\\j\neq i}}(g_{ij} - \bar{g_i})^2.
\end{aligned}
\end{equation}

\subsection{Interleaved Training}

Negative log-likelihood (NLL) has be shown to greatly overfit to ECE of the data it is trained on. Thus, training for calibration using the same data for the NLL has limited effect on reducing the calibration error of the model. \citet{softcal} proposed \textit{interleaved training} where they split the train set into two subsets - one is used to optimize the NLL and the other is used to optimize the calibration objective. Following this framework, let $\mathcal{D}_{train}$ denote the entire train set. We separate the train set into two subsets - $\mathcal{D}'_{train}$ and $\mathcal{D}'_{cal}$. The joint training of ESD with NLL becomes,
\begin{equation}
\underset{\theta}{\min} \ \text{NLL}(\mathcal{D}'_{train},\theta)+\lambda \cdot \text{ESD}(\mathcal{D}'_{cal},\theta).
\end{equation}
With this training scheme, NLL is optimized on $\mathcal{D}'_{train}$ and ESD is optimized on $\mathcal{D}'_{cal}$. This way, \citet{softcal} showed we can avoid minimizing ECE that is already overfit to the train set.

\section{Experimental Setting}

\subsection{Datasets and Models}
%\paragraph{TV show Retrieval}
% Dataset 한문장 설명

%We evaluate the proposed ESD metric as a trainable calibration objective on both image classification and natural language inference tasks.

\paragraph{Image Classification} For image classification tasks we use the following datasets:
\begin{itemize}
\setlength\itemsep{0em}
\item MNIST \citep{mnist}: 54,000/6,000/10,000 images for train, validation, and test split was used. We resized the images to (32x32) before inputting to the network.
\item CIFAR10 \& CIFAR100 \citep{cifar10,cifar100}: 45,000/5,000/10,000 images for train, validation, and test split was used. Used random cropping of 32 with padding of 4. Normalized each RGB channel with mean of 0.5 and standard deviation of 0.5.
\item ImageNet100 \citep{imagenet}: A subset dataset from ImageNet Large Scale Visual Recognition Challenge 2012 with 100 classes. Since the labels of test sets are unavailable, we use the official validation set as the test set, and we dedicate 10\% of the training data as the validation set for the experiments (i.e., 117,000/13,000/5,000 split for train/val/test set).
\end{itemize}

%Our model is evaluated on the val\_1 split.
%
\paragraph{Natural Language Inference (NLI)} NLI is a task in Natural Language Processing where it involves classifying  the inference relation (entailment, contradiction, or neutral) between two texts \citep{nli}. For NLI tasks we use the following datasets:
\begin{itemize}
    \setlength\itemsep{0em}
    \item SNLI \citep{snli}: SNLI corpus is a collection of human-written English sentence pairs manually labeled for balanced classification with the labels entailment, contradiction, and neutral. The data consists of 550,152/10,000/10,000 sentence pairs for train/val/test set respectively. The max length of the input is set to 158.
    \item ANLI \citep{anli}: ANLI dataset is a large-scale NLI dataset, collected via an, adversarial human-and-model-in-the-loop procedure. The data consists of 162,865/3,200/3,200 sentence pairs for train/val/test set respectively. The max length of the input is set to 128.
    
\end{itemize}
%\subsection{Model} 
For the Image Classification datasets, we used Convolutional Neural Networks (CNNs). Specifically, we used LeNet5 \citep{lenet}, ResNet50, ResNet34, and ResNet18 \citep{resnet}, for MNIST, CIFAR10, CIFAR100, and ImageNet100, respectively. For the NLI datasets, we finetuned transformer based Pre-trained Language Models (PLMs). Specifically, we used Bert-base \citep{bert} and Roberta-base \citep{roberta}, for SNLI and ANLI, respectively.

\subsection{Experimental setup}
\label{experimental_setup}
We compare our Expected Squared Difference (ESD) to previously proposed trainable calibration objectives, MMCE and SB-ECE, on the datasets and models previously mentioned. For fair comparison, interleaved training has been used for all three calibration objectives. For MMCE, in which \citet{MMCE} proposed an unweighted and weighted version of the objective, we use the former to set the method to account for the overfitting problem mentioned in section \ref{overfitproblem} consistent to interleaved training. For the interleaved training settings, we held out 10\% of the train set to the calibration set. 
\begin{table*}[ht]
    \setlength{\tabcolsep}{5pt}
	\centering

	\begin{tabular}{l|l||c c c c c}
		\Xhline{3\arrayrulewidth}
		                 Dataset & \multirow{2}{*}{Loss Fn.}
		                  & \multirow{2}{*}{Acc.}
		                  & \multirow{2}{*}{ECE}
		                  & \begin{small}ECE\end{small}
		                  & \begin{small}Acc.\end{small}
		                  & \begin{small}ECE\end{small}
		                  \\ 
		                (Model) & 
		                  &  
		                  &
		                  &\begin{small}\emph{after} TS\end{small}
		                  %& \begin{small}(LSTM)\end{small}
		                  & \begin{small}\emph{after} VS\end{small}
		                  & \begin{small}\emph{after} VS\end{small}
		                  \\ 
		\Xhline{2\arrayrulewidth}

		 & \begin{small}NLL (baseline)\end{small} & \begin{small}98.8$\pm$0.034\end{small} &  \begin{small}0.91$\pm$0.080\end{small} & \begin{small}0.31$\pm$0.044\end{small} & \begin{small}98.7$\pm$0.058\end{small} & \begin{small}0.43$\pm$0.123\end{small}\\
		 \begin{small}MNIST\end{small} & \begin{small}+MMCE\end{small} & \begin{small}98.4$\pm$0.320\end{small} &  \begin{small}0.36$\pm$0.029\end{small} & \begin{small}0.33$\pm$0.068\end{small} & \begin{small}98.4$\pm$0.250\end{small} & \begin{small}0.32$\pm$0.055\end{small}\\
		 \begin{small}(LeNet5)\end{small}& \begin{small}+SB-ECE\end{small} & \begin{small}97.9$\pm$0.177\end{small} &  \begin{small}0.41$\pm$0.067\end{small} & \begin{small}0.45$\pm$0.103\end{small} & \begin{small}97.9$\pm$0.073\end{small} & \begin{small}0.39$\pm$0.088\end{small}\\
		 & \begin{small}\bf{+ESD (ours)}\end{small} & \begin{small}98.6$\pm$0.204\end{small} &  \begin{small}\textbf{0.30}$\pm$0.035\end{small} & \begin{small}\textbf{0.29}$\pm$0.030\end{small} & \begin{small}98.6$\pm$0.167\end{small} & \begin{small}\textbf{0.28}$\pm$0.071\end{small}\\	
        \hline

		 & \begin{small}NLL (baseline)\end{small} & \begin{small}92.9$\pm$0.159\end{small} &  \begin{small}5.49$\pm$0.105\end{small} & \begin{small}1.89$\pm$0.105\end{small} & \begin{small}92.3$\pm$0.625\end{small} & \begin{small}2.96$\pm$1.596\end{small}\\
		 \begin{small}CIFAR10\end{small} & \begin{small}+MMCE\end{small} & \begin{small}91.5$\pm$0.340\end{small} &  \begin{small}4.92$\pm$0.292\end{small} & \begin{small}1.92$\pm$0.274\end{small} & \begin{small}91.3$\pm$0.429\end{small} & \begin{small}2.52$\pm$0.586\end{small}\\
		 \begin{small}(Resnet50)\end{small}& \begin{small}+SB-ECE\end{small} & \begin{small}91.6$\pm$0.288\end{small} &  \begin{small}4.86$\pm$0.319\end{small} & \begin{small}1.63$\pm$0.300\end{small} & \begin{small}91.5$\pm$0.343\end{small} & \begin{small}1.88$\pm$0.393\end{small}\\
		 & \begin{small}\bf{+ESD (ours)}\end{small} & \begin{small}92.1$\pm$0.141\end{small} &  \begin{small}\textbf{3.08}$\pm$0.692\end{small} & \begin{small}\textbf{1.60}$\pm$0.089\end{small} & \begin{small}92.1$\pm$0.314\end{small} & \begin{small}\textbf{1.61}$\pm$0.287\end{small}\\	
        \hline

		 & \begin{small}NLL (baseline)\end{small} & \begin{small}68.4$\pm$0.491\end{small} &  \begin{small}23.8$\pm$0.403\end{small} & \begin{small}5.74$\pm$0.306\end{small} & \begin{small}67.3$\pm$0.551\end{small} & \begin{small}9.78$\pm$0.640\end{small}\\
		 \begin{small}CIFAR100\end{small} & \begin{small}+MMCE\end{small} & \begin{small}66.5$\pm$0.644\end{small} &  \begin{small}13.9$\pm$0.545\end{small} & \begin{small}4.91$\pm$0.457\end{small} & \begin{small}66.3$\pm$0.773\end{small} & \begin{small}4.59$\pm$1.575\end{small}\\
		 \begin{small}(Resnet34)\end{small}& \begin{small}+SB-ECE\end{small} & \begin{small}67.3$\pm$0.367\end{small} &  \begin{small}14.7$\pm$1.370\end{small} & \begin{small}4.94$\pm$0.240\end{small} & \begin{small}66.4$\pm$0.369\end{small} & \begin{small}4.57$\pm$1.253\end{small}\\
		 & \begin{small}\bf{+ESD (ours)}\end{small} & \begin{small}67.4$\pm$0.356\end{small} &  \begin{small}\textbf{13.6}$\pm$0.950\end{small} & \begin{small}\textbf{4.85}$\pm$0.390\end{small} & \begin{small}67.1$\pm$0.496\end{small} & \begin{small}\textbf{4.28}$\pm$1.396\end{small}\\	
        \hline

		 & \begin{small}NLL (baseline)\end{small} & \begin{small}75.8$\pm$0.397\end{small} &  \begin{small}10.3$\pm$0.686\end{small} & \begin{small}2.51$\pm$0.378\end{small} & \begin{small}75.9$\pm$0.595\end{small} & \begin{small}2.86$\pm$0.459\end{small}\\
		 \begin{small}ImageNet100\end{small} & \begin{small}+MMCE\end{small} & \begin{small}74.3$\pm$0.248\end{small} &  \begin{small}3.83$\pm$0.644\end{small} & \begin{small}1.94$\pm$0.262\end{small} & \begin{small}74.5$\pm$0.444\end{small} & \begin{small}2.29$\pm$0.227\end{small}\\
		 \begin{small}(Resnet18)\end{small}& \begin{small}+SB-ECE\end{small} & \begin{small}74.4$\pm$0.596\end{small} &  \begin{small}4.28$\pm$0.318\end{small} & \begin{small}2.06$\pm$0.260\end{small} & \begin{small}74.7$\pm$0.472\end{small} & \begin{small}2.13$\pm$0.198\end{small}\\
		 & \begin{small}\bf{+ESD (ours)}\end{small} & \begin{small}74.6$\pm$0.320\end{small} &  \begin{small}\textbf{1.80}$\pm$0.262\end{small} & \begin{small}\textbf{1.72}$\pm$0.350\end{small} & \begin{small}74.8$\pm$0.436\end{small} & \begin{small}\textbf{1.87}$\pm$0.266\end{small}\\	
        \hline
        
		 & \begin{small}NLL (baseline)\end{small} & \begin{small}90.2$\pm$0.434\end{small} &  \begin{small}4.20$\pm$0.420\end{small} & \begin{small}1.04$\pm$0.112\end{small} & \begin{small}90.1$\pm$0.397\end{small} & \begin{small}0.96$\pm$0.099\end{small}\\
		 \begin{small}SNLI\end{small} & \begin{small}+MMCE\end{small} & \begin{small}89.4$\pm$0.491\end{small} &  \begin{small}1.11$\pm$0.181\end{small} & \begin{small}1.01$\pm$0.178\end{small} & \begin{small}89.5$\pm$0.588\end{small} & \begin{small}1.08$\pm$0.170\end{small}\\
		 \begin{small}(Bert-base)\end{small}& \begin{small}+SB-ECE\end{small} & \begin{small}89.1$\pm$0.668\end{small} &  \begin{small}1.82$\pm$0.301\end{small} & \begin{small}0.99$\pm$0.180\end{small} & \begin{small}89.2$\pm$0.765\end{small} & \begin{small}0.90$\pm$0.109\end{small}\\
		 & \begin{small}\bf{+ESD (ours)}\end{small} & \begin{small}89.3$\pm$0.536\end{small} &  \begin{small}\textbf{0.98}$\pm$0.165\end{small} & \begin{small}\textbf{0.73}$\pm$0.192\end{small} & \begin{small}89.4$\pm$0.583\end{small} & \begin{small}\textbf{0.61}$\pm$0.126\end{small}\\	
        \hline

		 & \begin{small}NLL (baseline)\end{small} & \begin{small}49.4$\pm$0.323\end{small} &  \begin{small}35.9$\pm$0.505\end{small} & \begin{small}4.16$\pm$0.445\end{small} & \begin{small}48.4$\pm$0.492\end{small} & \begin{small}5.10$\pm$0.657\end{small}\\
		 \begin{small}ANLI\end{small} & \begin{small}+MMCE\end{small} & \begin{small}49.0$\pm$0.471\end{small} &  \begin{small}31.2$\pm$0.850\end{small} & \begin{small}3.71$\pm$0.175\end{small} & \begin{small}47.7$\pm$0.429\end{small} & \begin{small}4.79$\pm$0.693\end{small}\\
		 \begin{small}(Roberta-base)\end{small}& \begin{small}+SB-ECE\end{small} & \begin{small}48.5$\pm$0.481\end{small} &  \begin{small}33.9$\pm$1.378\end{small} & \begin{small}3.98$\pm$0.733\end{small} & \begin{small}47.3$\pm$0.281\end{small} & \begin{small}5.10$\pm$0.119\end{small}\\
		 & \begin{small}\bf{+ESD (ours)}\end{small} & \begin{small}48.0$\pm$0.451\end{small} &  \begin{small}\textbf{28.8}$\pm$0.543\end{small} & \begin{small}\textbf{3.49}$\pm$0.373\end{small} & \begin{small}47.1$\pm$0.429\end{small} & \begin{small}\textbf{4.42}$\pm$1.010\end{small}\\	
        \hline

	\end{tabular}
	\caption{Average accuracy (\%) and ECE (\%) (with std. across 5 trials) for baseline, MMCE, SB-ECE, ESD after training and after post-processing with temperature scaling (TS) or vector scaling (VS). }
	\label{tab:main}
\end{table*}
The regularizer hyperparameter $\lambda$ for weighting the calibration measure with respect to NLL is chosen via fixed grid search \footnote{For $\lambda$, we search for [0.2, 0.4, 0.6, 0.8, 1.0, 2.0, 3.0, ... 10.0] (Appendix \ref{lambda choice}).}. For measuring calibration error, we use ECE with 20 equally sized bins. For the image classification tasks we use AdamW \citep{adamw} optimizer with $10^{-3}$ learning rate and $10^{-2}$ weight decay for 250 epochs, except for ImageNet100, in which case we used $10^{-4}$ weight decay for 90 epochs. For the NLI tasks, we use AdamW optimizer with $10^{-5}$ learning rate and $10^{-2}$ weight decay for 15 epochs. For both tasks, we use a batch size of 512. 

The internal hyperparameters within MMCE ($\phi$) and SB-ECE ($M$, $T$) were sequentially optimized \footnote{For $\phi$, we search [0.2, 0.4, 0.6, 0.8]. For $T$, we search [0.0001, 0.001, 0.01, 0.1].} following the search for optimal $\lambda$. Following the original experimental setting of SB-ECE by \citet{softcal}, we fix the hyperparameter $M$ to 15. Similar to the model selection criterion utilized by \citet{softcal}, we look at the accuracy and the ECE of all possible hyperparameter configurations in the grid. We choose the lowest ECE while giving up less than $1.5\%$ accuracy relative to baseline accuracy on the validation set. All experiments were done using NVIDIA Quadro RTX 8000 and NVIDIA RTX A6000.

%\paragraph{Training Details.} 
%Our model is trained on NVIDIA Quadro RTX 8000 (48GB of memory) GPU. The dimensoin of hidden layer is $d = 768$ and we use AdamW optimizer \cite{loshchilov2017decoupled} with  a learning rate of $2.5e-5$ weight decay of 0.01 to train the model. The training hyperparameters in Equation \ref{eq:31} are $\alpha = 4, \beta = 2$, and $\gamma = 0.1$.

\section{Experimental Result}
In Table \ref{tab:main}, we report the accuracy and ECE for the models before and after post-processing (i.e., temperature scaling and vector scaling) of various datasets and models. Compared to the baseline or other trainable calibration objective loss (MMCE and SB-ECE), jointly training with ESD as the secondary loss consistently results in a better-calibrated network for all the datasets and models with around 1\% degradation of accuracy. Moreover, we observe that applying post-processing (i.e., temperature scaling (TS) and vector scaling (VS)) after training with a calibration objective loss generally results in better-calibrated models over baseline post-processing, in which case ESD still outperforms other methods. Comparing the calibration outcomes in temperature scaling with vector scaling for models trained with ESD, vector scaling worked comparable if not better as a post-processing method, except for ANLI, with minimal impact on the accuracy for all datasets. Additionally, \citet{datashift2} demonstrated that postprocessing methods, such as temperature scaling, are not effective in achieving good calibration under distribution shifts. Building on this research, \citet{softcal} have provided evidence that incorporating calibration during model training can enhance a model's ability to maintain calibration under distribution shifts. This finding has been replicated in Appendix \ref{distributionshiftresult}.

\begin{figure}[t]
	\centering
	\includegraphics[width=0.9\linewidth]{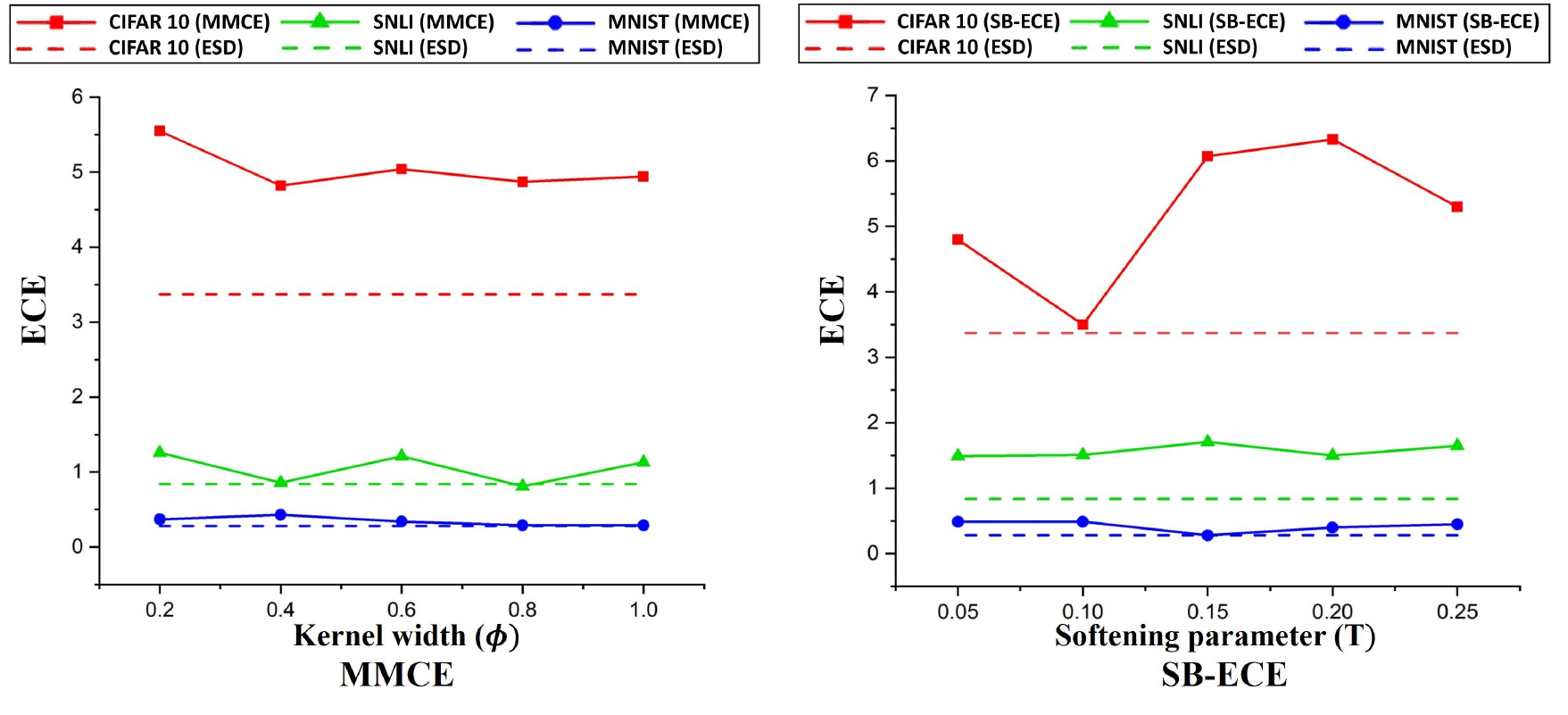}
	\caption{ECE performance curve of MMCE (left) and SB-ECE (right) with respect to their varying internal hyperparameters on MNIST, CIFAR10, SNLI datasets. 
	%Since the optimal hyperparameter changes for different models and datasets, they require a tuning process. On the other hand, our proposed ESD not only outperforms MMCE and SB-ECE but does not have an internal hyperparameter to tune for.
	}
	\label{sensitivity}
\end{figure}

We show in Figure \ref{sensitivity} that across different models and datasets the calibration performance of MMCE and SB-ECE is sensitive to the internal hyperparameter. This shows the importance of hyperparameter tuning in these methods. On the other hand, ESD does not only outperforms MMCE and SB-ECE in terms of ECE but does not have an internal hyperparameter to tune for.

\begin{figure}[t]
	\centering
	\includegraphics[width=0.85\linewidth]{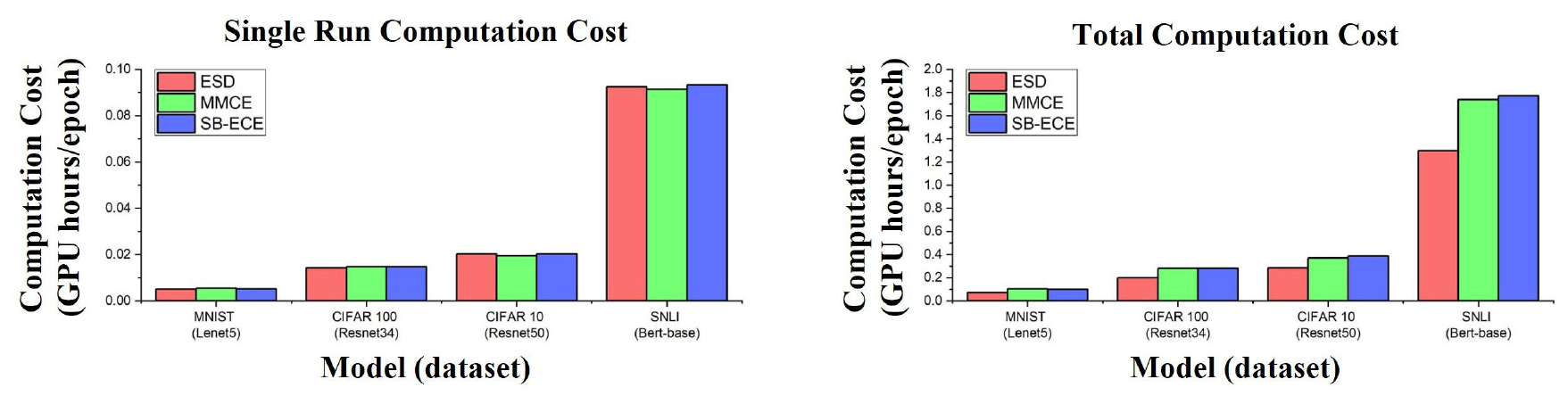}
	\caption{Computational cost of single-run training (left) and total cost considering hyperparameter tuning (right). The x-axis in both cases are in the order of increasing model complexity.}
	\label{fig:computation}
\end{figure}

\section{Ablation Study}

\subsection{Computational Cost of Hyperparameter Seach}

We investigate the computational cost required to train a model with ESD compared to MMCE and SB-ECE. As depicted in Figure \ref{fig:computation}, the computational cost required to train a model for a single run remains nearly identical across different models and datasets. However, considering the need for tuning additional hyperparameters within MMCE and SB-ECE, the discrepancy in total computational cost between ESD and tuning-required calibration objective losses becomes more prominent as the model complexity and dataset size increases.

\subsection{Batch Size Experiments}
Table \ref{tab:batch} shows that ESD is robust to varying batch sizes which could be due to the fact that it is an unbiased and consistent estimator. Regardless of the batch size, being unbiased guarantees that the average of the gradient vectors of ESD equals the gradient of the true estimate (Appendix \ref{gradient}).

\begin{table*}[t]
	\centering

	\begin{tabular}{l|l||c c c c c}
		\Xhline{3\arrayrulewidth}
		                 Dataset & \multirow{2}{*}{Loss Fn.}
		                  & \multirow{2}{*}{Acc.}
		                  & \multirow{2}{*}{ECE}
		                  & \begin{small}ECE\end{small}
		                  & \begin{small}Acc.\end{small}
		                  & \begin{small}ECE\end{small}
		                  \\ 
		                (Model) & 
		                  &  
		                  &
		                  &\begin{small}\emph{after} TS\end{small}
		                  %& \begin{small}(LSTM)\end{small}
		                  & \begin{small}\emph{after} VS\end{small}
		                  & \begin{small}\emph{after} VS\end{small}
		                  \\ 
		\Xhline{2\arrayrulewidth}

		 & \begin{small}NLL (b512)\end{small} & \begin{small}92.9$\pm$0.159\end{small} &  \begin{small}5.49$\pm$0.105\end{small} & \begin{small}1.89$\pm$0.105\end{small} & \begin{small}92.3$\pm$0.625\end{small} & \begin{small}2.96$\pm$1.596\end{small}\\
		 \begin{small}CIFAR10\end{small} & \begin{small}+ESD (b512)\end{small} & \begin{small}92.1$\pm$0.141\end{small} &  \begin{small}3.08$\pm$0.692\end{small} & \begin{small}1.60$\pm$0.089\end{small} & \begin{small}92.1$\pm$0.314\end{small} & \begin{small}1.61$\pm$0.287\end{small}\\
		 \begin{small}(Resnet50)\end{small}& \begin{small}+ESD (b256)\end{small} & \begin{small}91.5$\pm$0.137\end{small} &  \begin{small}3.15$\pm$0.587\end{small} & \begin{small}1.61$\pm$0.073\end{small} & \begin{small}91.9$\pm$0.573\end{small} & \begin{small}1.60$\pm$0.082\end{small}\\
		 & \begin{small}+ESD (b128)\end{small} & \begin{small}91.5$\pm$0.519\end{small} &  \begin{small}2.93$\pm$0.539\end{small} & \begin{small}1.70$\pm$0.600\end{small} & \begin{small}91.7$\pm$0.246\end{small} & \begin{small}2.20$\pm$0.584\end{small}\\	
		 & \begin{small}+ESD (b64)\end{small} & \begin{small}91.3$\pm$0.753\end{small} &  \begin{small}3.91$\pm$0.847\end{small} & \begin{small}1.87$\pm$0.132\end{small} & \begin{small}91.5$\pm$0.257\end{small} & \begin{small}2.65$\pm$1.48\end{small}\\       
        \hline

		 & \begin{small}NLL (b512)\end{small} & \begin{small}90.2$\pm$0.434\end{small} &  \begin{small}4.20$\pm$0.420\end{small} & \begin{small}1.04$\pm$0.112\end{small} & \begin{small}90.1$\pm$0.397\end{small} & \begin{small}0.96$\pm$0.099\end{small}\\
		 \begin{small}SNLI\end{small}& \begin{small}+ESD (b512)\end{small} & \begin{small}89.3$\pm$0.536\end{small} &  \begin{small}0.98$\pm$0.165\end{small} & \begin{small}0.73$\pm$0.192\end{small} & \begin{small}89.4$\pm$0.583\end{small} & \begin{small}0.61$\pm$0.126\end{small}\\
		 \begin{small}(Bert-base)\end{small}& \begin{small}+ESD (b256)\end{small} & \begin{small}89.9$\pm$0.314\end{small} &  \begin{small}1.11$\pm$0.236\end{small} & \begin{small}0.76$\pm$0.241\end{small} & \begin{small}88.9$\pm$0.146\end{small} & \begin{small}0.62$\pm$0.147\end{small}\\
		 
		 & \begin{small}+ESD (b128)\end{small} & \begin{small}89.8$\pm$0.435\end{small} &  \begin{small}0.99$\pm$0.244\end{small} & \begin{small}0.68$\pm$0.190\end{small} & \begin{small}89.6$\pm$0.490\end{small} & \begin{small}0.58$\pm$0.187\end{small} \\
		 & \begin{small}+ESD (b64)\end{small} & \begin{small}89.1$\pm$0.660\end{small} &  \begin{small}1.06$\pm$0.285\end{small} & \begin{small}0.83$\pm$0.260\end{small} & \begin{small}88.8$\pm$0.672\end{small} & \begin{small}0.81$\pm$0.153\end{small} \\
        \hline

	\end{tabular}
	\caption{Average accuracy and ECE (with std. across 5 trials) for ESD after training with batch sizes 64, 128, 256, and 512.}
	\label{tab:batch}
\end{table*}

\subsection{Are Indicator Functions Trainable?}
\label{ece}
Recent papers, \citet{MMCE} and \citet{softcal}, have suggested that ECE is not suitable for training as a result of its high discontinuity due to binning, which can be seen as a form of an indicator function. However, the results from our method suggest that our measure was still able to train well despite the existence of indicator functions. In addition, previous measures also contain indicator functions in the form of \textit{argmax} function that introduces discontinuities but remains to be trainable. As such, this brings to rise the question of whether calibration measures with indicator functions can be used for training. To investigate this, we ran ECE as an auxiliary calibration loss on different batch sizes and observed its performance on CIFAR 100 (Table \ref{tab:batchECE}).

We found that ECE, contrary to previous belief, is trainable under large batch sizes and not trainable under small batch sizes while ours maintains good performance regardless of batch size. The poor performance of ECE under small batch size setting could potentially be attributed to the high bias present in such cases. From this, it seems to suggest that indicator functions do not seem to inhibit the training for calibration. 

\begin{table*}[ht]
	\centering
	 \begin{tabular}{l||c c }
		\Xhline{3\arrayrulewidth}
		                 \multirow{1}{*}{Loss Fn.}
		                  & \multirow{1}{*}{Acc.}
		                  & \multirow{1}{*}{ECE}
		                  \\ 
		\Xhline{2\arrayrulewidth}
        
		 \begin{small}NLL (b512)\end{small} & \begin{small}68.6$\pm$0.034\end{small} &  \begin{small}22.4$\pm$0.105\end{small}\\
 		 \begin{small}+ECE (b512)\end{small} & \begin{small}68.2$\pm$0.025\end{small} &  \begin{small}14.6$\pm$0.252\end{small}\\
		 \begin{small}+ECE (b256)\end{small} & \begin{small}67.9$\pm$0.057\end{small} &  \begin{small}18.1$\pm$0.552\end{small}\\
		 \begin{small}+ECE (b128)\end{small} & \begin{small}68.1$\pm$0.084\end{small} &  \begin{small}21.0$\pm$0.552\end{small}\\
        \hline
	    \end{tabular}
	\caption{Average accuracy and ECE (with std. across 5 trials) for CIFAR100 after training with ECE on batch sizes 128, 256, and 512.}
	\label{tab:batchECE}
\end{table*}

\section{Conclusions}
Motivated by the need for a tuning-free trainable calibration objective, we proposed Expected Squared Difference (ESD), which does not contain any internal hyperparameters. With extensive comparison with existing methods for calibration during training across various architectures (CNNs \& Transformers) and datasets (in vision \& NLP domains), we demonstrate that training with ESD provides the best-calibrated models compared to other methods with minor degradation in accuracy over baseline. Furthermore, the calibration of these models is further improved after post-processing. In addition, we demonstrate that ESD can be utilized in small batch settings while maintaining performance. More importantly, in contrast to previously proposed trainable calibration objectives, ESD does not contain any internal hyperparameters, which significantly reduces the total computational cost for training. This reduction in cost is more prominent as the complexity of the model and dataset increases, making ESD a more viable calibration objective option.
\newpage 
\section*{Acknowledgement}
This work was supported by Institute of Information \& communications Technology Planning \& Evaluation (IITP) grant funded by the Korea government(MSIT) (No.2022-0-00184, Development and Study of AI Technologies to Inexpensively Conform to Evolving Policy on Ethics), and Institute for Information \& communications Technology Promotion(IITP) grant funded by the Korea government(MSIT) (No. 2021-0-01381,
Development of Causal AI through Video Understanding and Reinforcement Learning, and Its Applications to Real Environments).

\bibliography{main}
\bibliographystyle{iclr2023_conference}

\newpage

\appendix
\section{Visual Intuition of Expected Squared Difference (ESD)}
\label{intuition}
\begin{figure}[ht] 
	\centering
	\includegraphics[width=.8\linewidth]{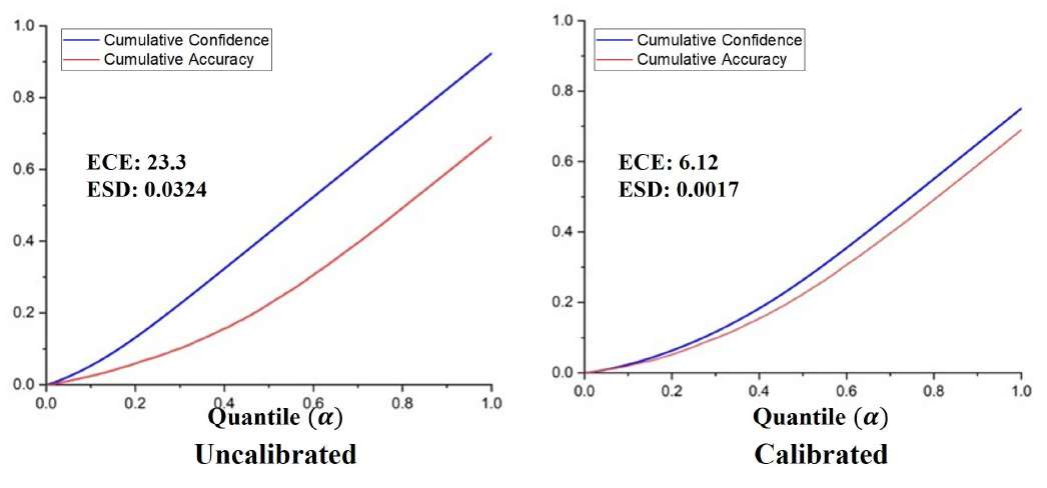}
	\caption{Visual intuition plot showing the cumulative confidence and cumulative accuracy with varying quantile scores of prediction confidence ($\alpha$) for an uncalibrated (left) and calibrated (right) network. The uncalibrated network was obtained by training Resnet34 on CIFAR100 with NLL, and the calibrated network was acquired by temperature scaling on the aforementioned trained network.}
	\label{fig:intuition}
\end{figure}

From Eq. (\ref{eq:7}),
\begin{equation} 
\begin{aligned}
d_k(\alpha) &= \left\vert \int^\alpha_{0}\mathbb{P}(Y_k = 1, Z_k = z_k) - z_k\mathbb{P}(Z_k = z_k)dz_k \right\vert\\
            &= |\mathbb{E}_{Z_k,Y_k}[I(Z_{k}\leq \alpha)(I(Y_k=1)-Z_{k})]|. \notag
\end{aligned}
\end{equation}
This can be viewed as the difference between two quantities:
\begin{equation}
\begin{aligned}
\text{\bf{Cumulative Accuracy}}    &=  \int^\alpha_{0}\mathbb{P}(Y_k = 1, Z_k = z_k)dz_k\\
                                &= \mathbb{E}_{Z_k,Y_k}[I(Z_{k}\leq \alpha, Y_k=1)]\\
                                &\approx \frac{1}{N}\sum_{i=1}^N I(z_{k,i}\leq \alpha, Y_{k,i}=1).\\
\text{\bf{Cumulative Confidence}}      &=  \int^\alpha_{0} z_k\mathbb{P}(Z_k = z_k)dz_k \\
                                &= \mathbb{E}_{Z_k,Y_k}[Z_{k}I(Z_{k}\leq \alpha)]\\
                                &\approx \frac{1}{N}\sum_{i=1}^N z_{k,i}I(z_{k,i}\leq \alpha).\\  \notag
\end{aligned}
\end{equation}
Due to the close relationship between $d_k(\alpha)$ and the calibration of a neural network, the average squared difference ($\mathbb{E}_{Z'_k}[d_k(Z'_k)^2]$) between the cumulative accuracy and confidence closely corresponds with the calibration of a network (Figure \ref{fig:intuition}). That is, the average squared difference between the cumulative accuracy and confidence is larger for an uncalibrated network compared to a calibrated network. Jointly training with our proposed Expected Squared Difference (ESD) as an auxiliary loss tries to minimize this squared difference between the two curves during training on average, thus achieving a better calibrated model.  %The below figure depicts this behaviour.

\section{Proof that ESD is an unbiased and consistent estimator}
\label{unbias}
In Theorem 2, we prove that ESD is an unbiased estimator, that is taking the expectation of ESD is equal to the true Expected Squared Difference.\\
{\bf Theorem 2} 
{\it ESD is an unbiased estimator, i.e. $\mathbb{E}_{\substack{\boldsymbol{Z_k}},\boldsymbol{Y_k}}[ESD]$ = $\mathbb{E}_{Z'_k}[d_k^2(Z'_k)]]$ where $d_k(Z'_k) = |\mathbb{E}_{Z_k,Y_k}[I(Z_{k}\leq Z'_{k})(I(Y_k=1)-Z_{k})]|$.}
\begin{proof}
Let $\boldsymbol{Z_k}$=($Z_{k,1}$,...,$Z_{k,n}$), $\boldsymbol{Y_k}$ = ($Y_{k,1}$,...,$Y_{k,n}$) and $G_i = \bar{g_i}^2-\frac{S_{g_i}^2}{N-1} $. We have that
\begin{equation}
\begin{aligned}
\mathbb{E}_{\substack{\boldsymbol{Z_k}},\boldsymbol{Y_k}} [ESD]  &= \frac{1}{N}\sum_{i=1}^N \mathbb{E}_{\substack{\boldsymbol{Z_k}},\boldsymbol{Y_k}}[G_i] \hspace{5mm}\text{(by linearity of expectation)} \\
	 &= \frac{1}{N}\sum_{i=1}^N \mathbb{E}_{Z_{k,i},Y_{k,i}}[\mu^2_i] \hspace{5mm}\text{(by lemma 2.1)}\\  
  &=  \frac{1}{N}\sum_{i=1}^N \mathbb{E}_{Z'_k}[d_k^2(Z'_k)] \hspace{5mm}\text{(since $\mathbb{E}_{Z_{k,i},Y_{k,i}}[\mu^2_i] = \mathbb{E}_{Z'_k}[d_k^2(Z'_k)]$}\\
  &= \mathbb{E}_{Z'_k}[d_k^2(Z'_k)].
 \notag
\end{aligned}
\end{equation} 
\end{proof}

{\bf Lemma 2.1 }{\it $\mathbb{E}_{\boldsymbol{Z_{k,-i}},\boldsymbol{Y_{k,-i}}}[G_i] = \mu^2_i \text{ where } \mu_i = \mathbb{E}_{Z_k,Y_k}[I(Z_{k}\leq Z_{k,i})(I(Y_k=1)-z_{k})].$}
\begin{proof}
 Since samples are i.i.d., for a fixed $i$, it holds that:
\begin{equation}
\begin{aligned}
  \mathbb{E}_{\boldsymbol{Z_{k,-i}},\boldsymbol{Y_{k,-i}}}[\bar{g_i}] &= \mu_i\\
   \mathbb{E}_{\boldsymbol{Z_{k,-i}},\boldsymbol{Y_{k,-i}}}[\frac{S^2_{g_i}}{N-1}] &= \frac{\sigma_i^2}{N-1} \hspace{5mm}\text{(where }\text{\text{Var}}[g_i] = \sigma^2_i\text{)}\\
  \text{\text{Var}}[\bar{g_i}]    &= \mathbb{E}_{\boldsymbol{Z_{k,-i}},\boldsymbol{Y_{k,-i}}}[\bar{g_i}^2]-\mu_i^2.
\end{aligned}
\end{equation}
 Shifting variables around, we get
\begin{equation}
\begin{aligned}
	\mu_i^2                                  &= \mathbb{E}_{\boldsymbol{Z_{k,-i}},\boldsymbol{Y_{k,-i}}}[\bar{g_i}^2]-\text{\text{Var}}[\bar{g_i}]
 \\
 &= \mathbb{E}_{\boldsymbol{Z_{k,-i}},\boldsymbol{Y_{k,-i}}} \left[\bar{g_i}^2 - \frac{S^2_{g_i}}{N-1} \right] \hspace{5mm}\text{(since }\text{\text{Var}}[\bar{g_i}] = \frac{\sigma^2_i}{N-1}\text{)} % = \mathbb{E}_{\boldsymbol{Z_{k,-i}},\boldsymbol{Y_{k,-i}}}[\frac{S^2_{\bar{g_i}}}{N-1}] 
 \\		                                    &=\mathbb{E}_{\boldsymbol{Z_{k,-i}},\boldsymbol{Y_{k,-i}}}[G_i].
 \notag 
 \\
\end{aligned}
\end{equation}
\end{proof}

In Theorem 3, we prove that ESD is a consistent estimator, which means that ESD converges in probability to the true Expected Squared Difference.\\
{\bf Theorem 3}
{\it ESD is a consistent estimator, i.e. ESD $\xrightarrow{\mathbb{P}}$ $\mathbb{E}_{Z'_k}[ d_k^2(Z'_k)].$}
\begin{proof}
Since ESD is an unbiased estimator, it is sufficient to prove $\lim_{n\to\infty}\text{\text{Var}}[ESD] = 0$.
For simplicity, we use where $\sum_{i,j=1}^N  = \sum_{i=1}^N \sum_{j=1}^N $ and 
$\mathbb{E}_{\boldsymbol{Z_k},\boldsymbol{Y_k}}[G_i]  = \mathbb{E}_{Z'_k}[d_k^2(Z'_k)]\text{ } \forall i,$
\begin{equation}
\begin{aligned}
\text{Var}[ESD] &= \frac{1}{N^2}\smashoperator[l]{\sum_{i=1}^N} \text{Var}[G_i] + \frac{1}{N^2}\sum_{\substack{i,j=1\\ i\neq j}}^N \text{Cov}[G_i,G_j].
\hspace{5mm}\\
\rlap{\text{Since }$\mathbb{E}_{\boldsymbol{Z_k},\boldsymbol{Y_k}}[G_i]  = \mathbb{E}_{Z'_k}[d_k^2(Z'_k)]\text{ } \forall i,$}\hspace{4.25em}\\
		 &= \frac{1}{N^2}\smashoperator[l]{\sum_{i=1}^N} \text{Var}[G_i] + \frac{1}{N^2}\sum_{\substack{i,j=1\\ i\neq j}}^N (\mathbb{E}_{\boldsymbol{Z_k},\boldsymbol{Y_k}}[G_iG_j]-\mathbb{E}^2_{Z'_k}[d_k^2(Z'_k)])\hspace{5mm}\\
		 &\leq \frac{1}{N^2}\smashoperator[l]{\sum_{i=1}^N} 2 + \frac{1}{N^2}\sum_{\substack{i,j=1\\ i\neq j}}^N (\mathbb{E}_{\boldsymbol{Z_k},\boldsymbol{Y_k}}[\bar{g_i}^2\bar{g_j}^2] - \mathbb{E}^2_{Z'_k}[d_k^2(Z'_k)]) \hspace{5mm}\text{(by lemma 3.1 and 3.2)}   \\
       &\hspace{5mm}+ \frac{1}{N^2}\smashoperator[l]{\sum_{\substack{i,j=1\\ i\neq j}}^N}  \frac{4}{(N-2)^2}    \\
	    &= \frac{2}{N} + \frac{4N(N-1)}{N^2(N-2)^2} 
      + \frac{N(N-1)}{N^2}  (\mathbb{E}_{\boldsymbol{Z_k},\boldsymbol{Y_k}}[\bar{g_1}^2\bar{g_2}^2] - \mathbb{E}^2_{Z'_k}[d_k^2(Z'_k)]).\hspace{1mm} \text{(by lemma 3.3)}\\
   \notag 
\end{aligned}
\end{equation}
Thus, by using lemma 3.4 and the non-negativeness of variance,
\begin{equation}
\begin{aligned}
0 \le \lim_{n\to\infty}\text{Var}[ESD] &\leq \lim_{n\to\infty}\mathbb{E}_{\boldsymbol{Z_k},\boldsymbol{Y_k}}[\bar{g_1}^2\bar{g_2}^2] - \mathbb{E}^2_{Z'_k}[d_k^2(Z'_k)]  = 0. 
                          \notag \\
\end{aligned}
\end{equation}
Consequently, $\lim_{n\to\infty}\text{Var}[ESD] = 0.$\\
\end{proof}
{\bf Lemma 3.1}{\it \  $\text{Var}[G_i]\leq 2 < \infty$.}
\begin{proof}
\begin{equation}
\begin{aligned}
	\text{Var}[G_i] &= \mathbb{E}_{\boldsymbol{Z_{k}},\boldsymbol{Y_{k}}}[G_i^2] - \mathbb{E}_{\boldsymbol{Z_{k}},\boldsymbol{Y_{k}}}^2[G_i]\\
		&\leq \mathbb{E}_{\boldsymbol{Z_{k}},\boldsymbol{Y_{k}}}[G_i^2]\\
		 &= \mathbb{E}_{\boldsymbol{Z_{k}},\boldsymbol{Y_{k}}} \left[\bar{g_i}^2-2\frac{S_{g_i}^2}{N-1} + \left(\frac{S_{g_i}^2}{N-1} \right)^2 \right]\\
		&\leq \mathbb{E}_{\boldsymbol{Z_{k}},\boldsymbol{Y_{k}}}\left[\bar{g_i}^2+ \left(\frac{S_{g_i}^2}{N-1} \right)^2\right].
  \notag 
\end{aligned}
\end{equation}

By lemma 3.2, $ \mathbb{E}_{\boldsymbol{Z_{k}},\boldsymbol{Y_{k}}}[g_i^2]\leq 1$ and $ \mathbb{E}_{\boldsymbol{Z_{k}},\boldsymbol{Y_{k}}}\left[ \left(\frac{S_{g_i}^2}{N-1} \right)^2 \right] \leq \left(\frac{2}{N-2} \right)^2 \leq 1.$ Thus,\\
$\mathbb{E}_{\boldsymbol{Z_{k}},\boldsymbol{Y_{k}}}\left[\bar{g_i}^2+ \left(\frac{S_{g_i}^2}{N-1} \right)^2 \right] \le 2.$   
          
\end{proof}
{\bf Lemma 3.2}{\it \ $|\frac{S_{g_i}^2}{N-1}|\leq \frac{2}{N-2}$ and $|\bar{g_i}|\leq 1.$}
\begin{proof}
{
By the triangle inequality, 
$
|\bar{g_i}| = |\frac{1}{N-1}\sum\limits_{\substack{m=1\\ m\neq i}}^N g_{im}| \leq \frac{1}{N-1}\sum\limits_{\substack{m=1\\ m\neq i}}^N|g_{im}|\leq1
$, since $ \forall m$ $|g_{im}| \le 1.$ Furthermore, we have 
\begin{eqnarray*}
\left\vert \frac{S_{g_i}^2}{N-1} \right\vert    
&= & 
 \frac{1}{(N-1)^2}\sum_{\substack{m=1\\ m\neq i}}^N g_{im}^2 + \frac{1}{(N-1)^2(N-2)}(\sum_{\substack{m=1\\ m\neq i}}^N g_{im})^2
 \\
 &\leq&  
 \frac{1}{(N-1)^2}\sum_{\substack{m=1\\ m\neq i}}^N|g_{im}|^2 + \frac{1}{(N-1)^2(N-2)}(\sum_{\substack{m=1\\ m\neq i}}^N|g_{im}|)^2
 \\
 &\leq& 
 \frac{1}{(N-1)^2}\sum_{\substack{m=1\\ m\neq i}}^N1 + \frac{1}{(N-1)^2(N-2)}(\sum_{\substack{m=1\\ m\neq i}}^N 1)^2_. 
 \end{eqnarray*}
 The last inequality is by $\forall m$,  $|g_{im}|\leq 1$, and the direct calculation implies that $$ \left\vert \frac{S_{g_i}^2}{N-1} \right\vert  \leq  \frac{1}{N-1} + \frac{1}{N-2} \leq \frac{2}{N-2}_.$$
}

\end{proof}
{\bf Lemma 3.3 }{\it $\mathbb{E}_{\boldsymbol{Z_k},\boldsymbol{Y_k}}[\bar{g_i}^2\bar{g_j}^2] = \mathbb{E}_{\boldsymbol{Z_k},\boldsymbol{Y_k}}[\bar{g_1}^2\bar{g_2}^2]$ $\forall i,j$ where $i \neq j.$}
\begin{proof}
\begin{equation}
\begin{aligned}
	\sum_{\mathclap{\substack{m,n=1\\ \{m,n\} \neq \{i,j\}}}}^N  g_{im}g_{jn} \hspace{2mm} =\hspace{2mm}  &\smashoperator[l]{\sum_{\mathclap{\substack{m,n=1\\ \{m,n\} \neq \{i,j\} \text{, }m \neq n}} }^N}  g_{im}g_{jn} + \sum_{\mathclap{\substack{m,n=1\\ n \neq \{i,j\}}}}^N g_{ij}g_{jn} + \sum_{\mathclap{\substack{m,n=1\\ m \neq \{i,j\}}}}^N g_{im}g_{ji} && \\
	                                       + &\smashoperator[l]{\sum_{\mathclap{\substack{m,n=1\\ m \neq \{i,j\}}}}^N} g_{im}g_{jm} + g_{ij}g_{ji} \hspace{5mm}\text{(since these terms are disjoint terms)}\\
	 \overset{d}{=}  &\smashoperator[l]{\sum_{\mathclap{\substack{m',n'=1\\\{m',n'\} \neq \{1,2\} \text{, } m' \neq n'}}}^N}g_{1m'}g_{2n'} + \sum_{\mathclap{\substack{m',n'=1\\ n' \neq \{1,2\}}}}^N g_{12}g_{2n'} + \sum_{\mathclap{\substack{m',n'=1\\ m' \neq \{1,2\}}}}^N g_{1m'}g_{21} && \\
	                                       + &\smashoperator[l]{\sum_{\mathclap{\substack{m',n'=1\\ m' \neq \{1,2\}}}}^N} g_{1m'}g_{2m'} + g_{12}g_{21}\\
	                                   = &\smashoperator[l]{\sum_{\mathclap{\substack{m',n'=1\\ m' \neq 1, n' \neq 2}}}^N} g_{1m'}g_{2n'.} \\
	                                   \notag
	                                       \notag
\end{aligned}
\end{equation}
Proof of claim above:\\
As the summation can be divided into five distinct cases,\\
Let $i$ and $j$ be arbitrary such that $i\neq j$,
\begin{equation}
\begin{aligned}
\rlap{Case 1: $\{m,n\} \neq \{i, j\}$ and $m \neq n$,} \hspace{3em}\\
g_{im}g_{jn} &= I(Z_{k,m}\leq Z_{k,i})[I(Y_{k,m}=1)-Z_{k,m}]I(Z_{k,n}\leq Z_{k,j})[I(Y_{k,n}=1)-Z_{k,n}]\\
	&= h_1(Z_{k,i},Z_{k,m},Z_{k,j},Z_{k,n},Y_{k,m},Y_{k,n}) \\
	&\overset{d}{=} h_1(Z_{k,1},Z_{k,m'},Z_{k,2},Z_{k,n'},Y_{k,m'},Y_{k,n'}).  \hspace{2em}\text{where $\{m',n'\} \neq \{1, 2\}$ and $m' \neq n'$}\\
\rlap{Case 2: $m = j$ and $n \neq \{i, j\}$,} \hspace{3em}\\
g_{ij}g_{jn} &= I(Z_{k,j}\leq Z_{k,i})[I(Y_{k,j}=1)-Z_{k,j}] I(Z_{k,n}\leq Z_{k,j})[I(Y_{k,n}=1)-Z_{k,n}] \\
	&= h_2(Z_{k,i},Z_{k,j},Z_{k,n},Y_{k,j},Y_{k,n}) \\
	&\overset{d}{=} h_2(Z_{k,1},Z_{k,2},Z_{k,n'},Y_{k,2},Y_{k,n'}).\hspace{2em}\text{where  $m' = 2$ and $n' \neq \{1, 2\}$}\\
\rlap{Case 3: $n = i$ and $m \neq \{i,j\}$,} \hspace{3em}\\
g_{im}g_{ji} &= I(z_{k,m}\leq Z_{k,i})[I(Y_{k,m}=1)-Z_{k,m}]I(Z_{k,i}\leq Z_{k,j})[I(Y_{k,i}=1)-Z_{k,i}]\\
	&= h_3(Z_{k,i},Z_{k,m},Z_{k,j},Y_{k,m},Y_{k,i})\\
	&\overset{d}{=} h_3(Z_{k,1},Z_{k,m'},Z_{k,2},Y_{k,m'},Y_{k,1}).\hspace{2em}\text{where $n' = 1$ and $m' \neq \{1, 2\}$}\\
\rlap{Case 4: $m = n \neq \{i,j\}$,} \hspace{3em}\\
g_{im}g_{jm} &= I(Z_{k,m}\leq Z_{k,i})[I(Y_{k,m}=1)-Z_{k,m}]I(Z_{k,m}\leq Z_{k,j})[I(Y_{k,m}=1)-Z_{k,m}]\\
	&= h_4(Z_{k,i},Z_{k,m},Z_{k,j},Y_{k,m})\\
	&\overset{d}{=} h_4(Z_{k,1},Z_{k,m'},Z_{k,2},Y_{k,m'}). \hspace{2em}\text{where $m' = n'\neq \{1,2\}$}\\
\rlap{Case 5: $m = j$ and $n = i$,} \hspace{3em}\\
g_{ij}g_{ji} &= I(Z_{k,j}\leq Z_{k,i})[I(Y_{k,j}=1)-Z_{k,j}]I(Z_{k,i}\leq Z_{k,j})[I(Y_{k,i}=1)-Z_{k,i}\\
	&= h_5(Z_{k,i},Z_{k,j},Y_{k,i},Y_{k,j})  \\
	&\overset{d}{=} h_5(Z_{k,1},Z_{k2},Y_{k,1},Y_{k,2}). \hspace{2em}\text{where $m' = 2$  and $n'= 1$}\\
	\notag
\end{aligned}
\end{equation}
This follows from the fact that $Z_{k,i}$'s and $Y_{k,i}$'s are i.i.d. random variables.
Therefore,
\begin{equation}
\begin{aligned}
 (\sum_{\mathclap{\substack{m,n=1\\ m \neq i, n \neq j}}}^N g_{im}g_{jn})^2 &\overset{d}{=} (\sum_{\mathclap{\substack{m',n'=1\\ m' \neq 1, n' \neq 2}}}^N g_{1m'}g_{2n'})^2\\
\bar{g_i}^2\bar{g_j}^2 &\overset{d}{=} \bar{g_1}^2\bar{g_2}^2\\
 \mathbb{E}_{\boldsymbol{Z_k},\boldsymbol{Y_k}}[\bar{g_i}^2\bar{g_j}^2] &= \mathbb{E}_{\boldsymbol{Z_k},\boldsymbol{Y_k}}[\bar{g_1}^2\bar{g_2}^2].\\
\notag
\end{aligned}
\end{equation}

\end{proof}
{\bf Lemma 3.4}{\ \it $\lim_{N\to\infty}  \mathbb{E}_{\boldsymbol{Z_k},\boldsymbol{Y_k}}[\bar{g_1}^2\bar{g_2}^2] = \mathbb{E}^2_{Z'_k}[d_k^2(Z'_k)]$}.
\begin{proof}
For arbitrary $i,$ $\bar{g_i} \xrightarrow{\mathbb{P}} \mu_i$ by the strong law of large number. Thus, $\bar{g_1} \xrightarrow{\mathbb{P}} \mu_1$  and $\bar{g_2} \xrightarrow{\mathbb{P}} \mu_2$. Therefore, since marginal convergence in probability implies joint convergence in probability \citep{vaart_1998},
\begin{eqnarray} 
(\bar{g_1},\bar{g_2})
& \xrightarrow{\mathbb{P}} & (\mu_1,\mu_2).
\notag
\end{eqnarray} 
By continuous mapping theorem, since $f(x,y) = x^2y^2$ is a continuous function, \begin{eqnarray} 
\bar{g_1}^2\bar{g_2}^2 
& \xrightarrow{\mathbb{P}} & \mu_1^2\mu_2^2 .
\notag
\end{eqnarray}
Additionally, since $|\bar{g_1}^2\bar{g_2}^2|\leq 1 \  \forall N$, it is uniformly bounded and thus uniformly integrable. Combining this with the fact that it converges in probability,
\begin{eqnarray} 
\lim_{N\to\infty}\mathbb{E}_{\boldsymbol{Z_k},\boldsymbol{Y_k}}[\bar{g_1}^2\bar{g_2}^2] = \mathbb{E}_{\boldsymbol{Z_k},\boldsymbol{Y_k}}[\mu_1^2\mu_2^2].
\notag
\end{eqnarray}
Thus,
\begin{eqnarray}
\lim_{N\to\infty}\mathbb{E}_{\boldsymbol{Z_k},\boldsymbol{Y_k}}[\bar{g_1}^2\bar{g_2}^2] 
= \mathbb{E}_{\boldsymbol{Z_k},\boldsymbol{Y_k}}[\mu_1^2\mu_2^2] 
= 
\mathbb{E}_{\boldsymbol{Z_k},\boldsymbol{Y_k}}[\mu_1^2]E_{\boldsymbol{Z_k},\boldsymbol{Y_k}}[\mu_2^2] = \mathbb{E}^2_{Z'_k}[d_k^2(Z'_k)]. 
\notag
\end{eqnarray}

\end{proof}

\section{Expected Squared Difference (ESD) Pseudocode}
\label{algo}
In this section, we provide a pseudocode for calculating the Expected Squared Difference (ESD) for a given batch output.

\begin{algorithm}[h]
\SetAlgoLined
    \PyComment{n:number of samples in a mini-batch.}\\
    \PyComment{confidence:1D tensor of n elements containing max softmax outputs of each sample from a neural network.}\\
    \PyComment{correct:1D tensor of n elements containing 1/0 corresponding to the correctness of each sample.}\\
    \Pyoperator{def} ESD\_loss(n, confidence, correct): \\
    \Indp
    \PyComment{compute the difference between confidence and correctness.}\\
    diff = correct.float() - confidence\\

    \Pyoperator \\
    \PyComment{Prepare the split between inner and outer expectation estimation.}\\
    split = torch.ones(n,n) - torch.eye(n)\\
    \Pyoperator \\
    \PyComment{compute the inner expectation estimation.}\\
    confidence\_mat= confidence.expand(n,n)\\
    ineq = torch.le(confidence\_mat,confidence\_mat.T).float()\\
    diff\_mat = diff.view(1,n).expand(n,n)\\
    x\_mat = torch.mul(diff\_mat,ineq)*split\\
    mean\_row = torch.sum(x\_mat, dim = 1)/(n-1)\\
    x\_mat\_squared = torch.mul(x\_mat, x\_mat)\\
    var = 1/(n-2) * torch.sum(x\_mat\_squared,dim=1) - (n-1)/(n-2) * torch.mul(mean\_row,mean\_row)\\
    \Pyoperator \\
    \PyComment{compute the outer expectation estimation.}\\
    d\_k\_sq\_vector = torch.mul(mean\_row, mean\_row) - var/(n-1)\\
    ESD = torch.sum(d\_k\_sq\_vector)/n\\ 
    \Pyoperator{return} ESD \\
    \Indm
\caption{Pytorch-like Pseudocode: Expected Squared Difference (ESD)}
\label{algo:your-algo}
\end{algorithm}

\section{Unbiased estimators and its Gradient}
\label{gradient}
Since ESD is an unbiased estimator,
\begin{equation}
    \begin{aligned}
        \mathbb{E}_{\substack{\boldsymbol{Z_k}},\boldsymbol{Y_k}} [ESD] &= \mathbb{E}_{Z'_k}[d_k^2(Z'_k)]\\
        \nabla_{\theta} \mathbb{E}_{Z'_k}[d_k^2(Z'_k)] &= \nabla_{\theta} \mathbb{E}_{\substack{\boldsymbol{Z_k}},\boldsymbol{Y_k}} [ESD]\\
        &= \nabla_{\theta} \mathbb{E}_{\substack{\boldsymbol{X}},\boldsymbol{Y}} [ESD] \hspace{5mm}\text{(by law of unconscious statistician)}\\
        &= \mathbb{E}_{\substack{\boldsymbol{X}},\boldsymbol{Y}} [\nabla_{\theta}ESD]. \hspace{5mm}\text{(since ($\boldsymbol{X},\boldsymbol{Y}$) are independent of $\theta$)}
        \notag
    \end{aligned}
\end{equation}
Thus, the gradient of the unbiased estimator is on average the gradient of the desired metric.

\section{Additional Information on the Choice of Lambda Range}
\label{lambda choice}
\begin{figure}[ht]
	\centering
	\includegraphics[width=0.9\linewidth]{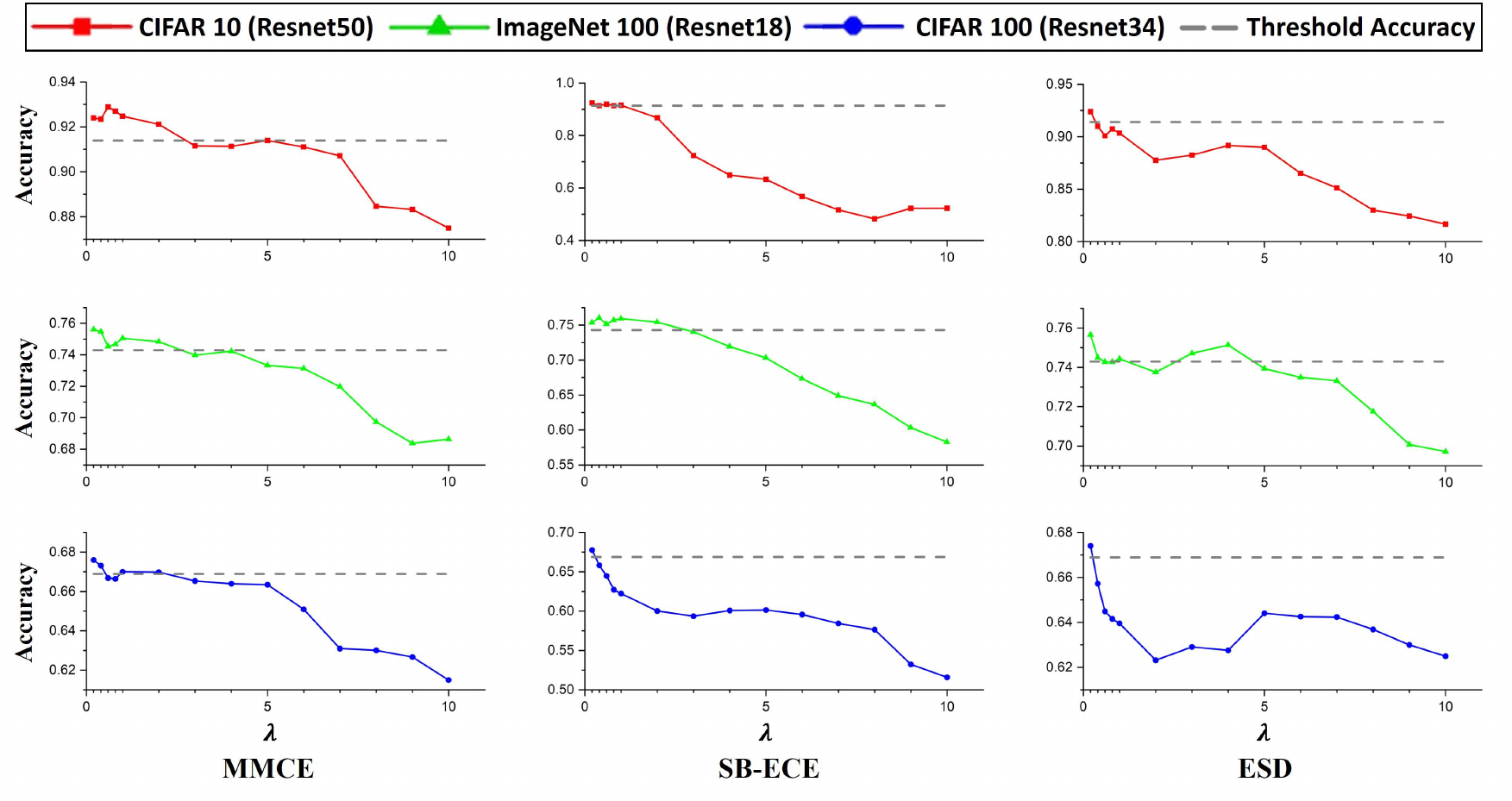}
	\caption{Accuracy plot with respect to varying values of $\lambda$ across different datasets and models trained with MMCE, SB-ECE, and ESD. The threshold accuracy represents the value 1.5\% below the baseline accuracy, which was used as the model selection criterion as stated in section \ref{experimental_setup}.}
	\label{fig:lambda_choice}
\end{figure}
To validate the choice of $\lambda$ grid range in our experimental settings, we plot the variations in accuracy with respect to increasing $\lambda$ for all methods (Figure \ref{fig:lambda_choice}). We see that our choice of $\lambda$ contains points within the acceptable range of accuracy (i.e., within 1.5\% degradation in accuracy compared to baseline). However, after a particular $\lambda$ value, which is different on a per method per dataset basis, the accuracy decreases consistently with increasing $\lambda$. As such, the grid chosen is suitable for the experiments conducted. 

\section{Trainable Calibration Measures Under Distribution Shift}
\begin{figure}[ht]
	\centering
	\includegraphics[width=1\linewidth]{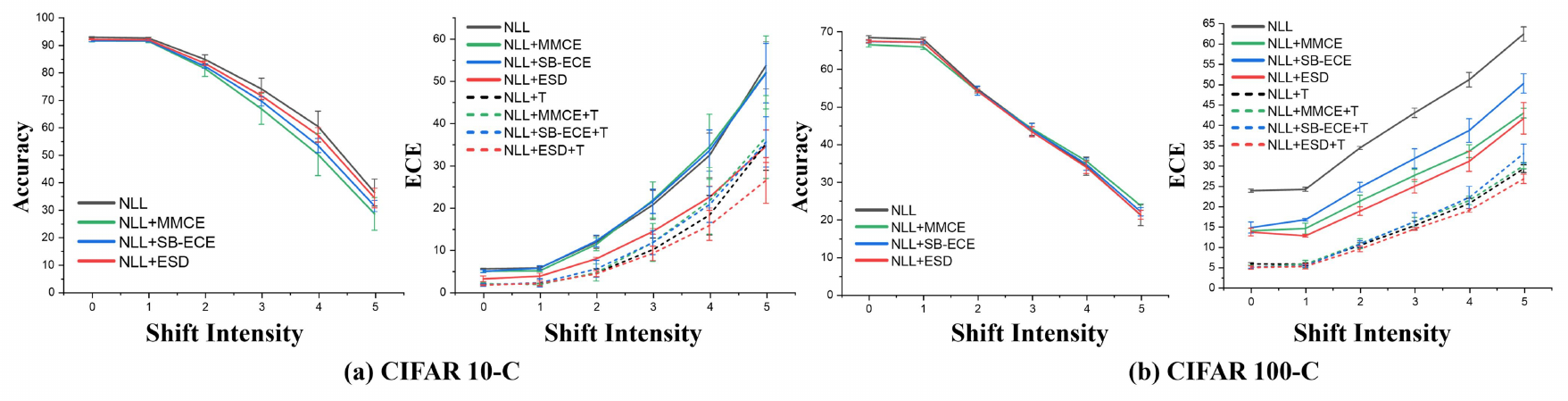}
	\caption{Accuracy and ECE plot for varying intensities of distribution shifts in (a) CIFAR 10-C and (b) CIFAR 100-C across models trained using NLL as well as those jointly trained with an auxiliary calibration objective (i.e., NLL+MMCE, NLL+SB-ECE, NLL+ESD) and their performance after post-processing with temperature scaling (i.e., NLL+T, NLL+MMCE+T, NLL+SB-ECE+T, NLL+ESD+T).}
	\label{fig:distribution shift}
\end{figure}
In Figure \ref{fig:distribution shift}, we evaluate the performance of calibration methods under distribution shift benchmark datasets, CIFAR 10-C and CIFAR 100-C, introduced in \citet{datashift}. %Figure \ref{fig:distribution shift}).
\label{distributionshiftresult}
For models jointly trained with an auxiliary calibration objective (NLL+MMCE, NLL+SB-ECE, NLL+ESD), we observe that they are more robust to distribution shifts when compared to those solely trained with NLL. In addition, prior work \citep{datashift2}, has shown that the calibration performance of a neural network after temperature scaling may be significantly reduced under distribution shifts. Our results on CIFAR 10-C and CIFAR 100-C imply a similar trend. Stacking calibration during training methods with temperature scaling (NLL+MMCE+T, NLL+SB-ECE+T, NLL+ESD+T), we observe that they perform comparable to temperature scaled models trained with NLL (NLL+T) with the exception of ESD, where it performs marginally better. As such, training with ESD could potentially improve a model's robustness after temperature scaling regarding distribution shifts and mitigate this issue. 
%Moreover, when stacked with temperature scaling, calibration during training methods such as SB-ECE and MMCE perform comparable to NLL after temperature scaling while ESD performs marginally better. As such, training with trainable calibration metrics such as ESD could potentially improve the robustness of temperature scaling with regards to distribution shift and potentially solve this issue.

%Stacking calibration during training methods with temperature scaling does not seem to help this issue as it still performs comparable if not slightly worse than NLL after temperature scaling. However, jointly training with ESD seems to improve the robustness 
\end{document}

%% file: math_commands.tex
\usepackage{amsmath,amsfonts,bm}

\def\eqref#1{equation~\ref{#1}}
\def\1{\bm{1}}

\DeclareMathAlphabet{\mathsfit}{\encodingdefault}{\sfdefault}{m}{sl}
\SetMathAlphabet{\mathsfit}{bold}{\encodingdefault}{\sfdefault}{bx}{n}

\DeclareMathOperator*{\argmin}{arg\,min}